\documentclass[12pt]{article}
\usepackage{xspace}
\usepackage{url,enumerate}
\usepackage{color,xcolor}
\usepackage{graphicx,epsfig,subfigure,wrapfig}
\usepackage{makeidx}
\usepackage{amsmath,amssymb,latexsym}
\usepackage{epstopdf}
\usepackage{graphicx}
\usepackage{subfigure}
\usepackage{algorithm}
\usepackage{algorithmic}
\usepackage{natbib}
\usepackage[titletoc]{appendix}
\RequirePackage[colorlinks,citecolor=blue,urlcolor=blue]{hyperref}

\newcommand{\w}{{\bf w}}
\newcommand{\noti}{{\backslash i}}
\newcommand{\Tr}{^{\rm T}}
\newcommand{\bphi}{\boldsymbol{\phi}}
\newcommand{\Beta}{\boldsymbol{\eta}}
\newcommand{\N}{{\cal N}}  % for normal density
\newcommand{\x}{{\bf x}}
\newcommand{\f}{{\bf f}}
\newcommand{\0}{{\bf 0}}
\newcommand{\balpha}{\boldsymbol{\alpha}}
\newcommand{\A}{{\bf A}}
\newcommand{\K}{{\bf K}}

\renewcommand*{\tt}{\tilde{t} \xspace}

\newcommand{\qnoti}{ q^{\noti} \xspace}

\newcommand{\qold}{q^{\textrm{old}} \xspace}
\newcommand{\bnu}{ \boldsymbol\nu \xspace}
\newcommand{\hatpi}{\hat{p}_i \xspace}
\newcommand{\lambdaib}{\tilde{\lambda_i}^{\noti} \xspace}
\newcommand{\hib}{\tilde{h_i}^{\noti} \xspace}
\newcommand{\mbi}{m_{i,b} \xspace}
\newcommand{\vbi}{v_{i,b} \xspace}
\newcommand{\alanc}[1]{}

\renewcommand{\t}{{\bf t}}
\renewcommand{\d}{\rm d}
\renewcommand{\a}{{\bf a}}
\newcommand{\email}[1]{\href{mailto:#1}{#1}}

\begin{document}

\title{Message passing with relaxed moment matching \footnote{This work was sponsored by the NSF grants IIS-0916443, IIS-1054903, ECCS-0941533, and CCF-0939370. All the authors gratefully acknowledge the support of the grants. Any opinions, findings, and conclusion
or recommendation expressed in this material are those of the author(s) and do not necessarily reflect the view of the funding
agencies or the U.S. government.}
}

\author{ {Yuan Qi} \\
        Departments of CS and Statistics \\
        Purdue University \\
        West Lafayette, IN 47907\\
        \email{alanqi@cs.purdue.edu}
         \and
        {Yandong Guo} \\
        School of ECE\\
        Purdue University \\
        West Lafayette, IN 47907 \\
        \email{guoy@purdue.edu}
}
\vspace{-0.02in}
\maketitle
\vspace{-0.02in}
\begin{abstract}
 Bayesian learning is often hampered by large computational expense.
As a powerful generalization of popular belief propagation,  expectation propagation (EP) efficiently approximates the exact Bayesian computation. Nevertheless, EP can be sensitive to outliers and suffer from divergence for difficult cases.  To address this issue, we propose a new approximate inference approach, relaxed expectation propagation (REP). It relaxes the moment matching requirement of expectation propagation by adding a relaxation factor into the KL minimization. We penalize this relaxation with a $l_1$ penalty. With this penalty, when two distributions in the relaxed KL divergence are similar, we obtain the exact moment matching; in the presence of outliers, the relaxation factor will used to relax the moment matching constraint.
 Based on this penalized KL minimization, REP is robust to outliers and can greatly improve the posterior approximation quality over EP.
  To examine the effectiveness of REP,  we apply it to Gaussian process classification, a task known to be suitable to EP.
Our classification results on synthetic and UCI benchmark datasets demonstrate significant improvement of REP over EP and Power EP---in terms of algorithmic stability, estimation accuracy and predictive performance.
\end{abstract}


\section{Introduction}

Bayesian learning provides a principled framework for modeling complex systems and making predictions.
A critical component of Bayesian learning is the computation of posterior distributions that represent estimation uncertainty. However, the exact computation is often so expensive that it has become a bottleneck for practical applications of Bayesian learning. To address this challenge, a variety of approximate inference methods has been developed to speed up the computation  \citep{Jaakkola:2000oz,Minka01Thesis,Opper:2005:ECA:1046920.1194917,Wainwright:2008}. As a representative approximate inference method, expectation propagation \citep{Minka01Thesis} generalizes the popular belief propagation algorithm, allows us to use structured approximations 
and handles both discrete and continuous posterior distributions. EP has been shown to significantly reduce computational cost while maintaining high approximation accuracy; for example, \citet{Kuss05} have demonstrated that, for Gaussian process (GP) classification, EP can provide accurate approximation to predictive posteriors. 

Despite its success in many applications, EP can be sensitive to outliers in observation and suffer from divergence when the exact distribution is not close to the approximating family used by EP.
This stems from the fact that EP approximates each factor in the model by a simpler form, known as messages, and iteratively refines the messages (See Section 2). Each message refinement is based on moment matching, which minimizes the Kullback-Leibler (KL) divergence between old and new beliefs.
The messages are refined in a distributed fashion---resulting in efficient inference on a graphical model.
But when the approximating family cannot fit the exact posterior well---such as 
in the presence of outliers---the message passing algorithm can suffer from divergence and give poor approximation quality.

We can force EP to converge by using the CCCP algorithm \citep{Yuille:2002rw,Heskes:2005sf}. But it is slower than the message passing updates. 
 Also, according to \citet{Minka01Thesis}, EP diverges for a good reason---indicating a poor approximating family or a poor energy function used by EP.

To address this issue, we propose a new approximate inference algorithm,  Relaxed Expectation Propagation (REP).  In REP, we introduce a relaxation factor $r$ in the KL minimization used by EP (See Section 3) and penalize this relaxation factor.  Because of this penalization, when the factor involved in the KL minimization is close to the current approximation, REP reduces to EP; when the factor is an outlier, the relaxation is used to stabilizing the message passing by relaxing the moment matching constraint. Regardless of the amount of outliers in data, REP converges in all of our experiments.  To better understand REP,  we also present the primal energy functions in Section~\ref{REP}. It differs from the EP energy function or the equivalent Bethe-like energy function \citep{Heskes:2005sf} by the use of relaxation factors.

To examine the performance of REP, in Section 5, we use it to train Gaussian process classification models for which EP is known to be a good choice for approximate inference \citep{Kuss05}.
In Section 7, we report experimental results on synthetic and UCI benchmark datasets, demonstrating that REP consistently outperforms EP and Power EP---in terms of algorithmic stability, estimation accuracy, and predictive performance.

\vspace{-0.15in}
\section{Background: Expectation Propagation}
Given observations ${\cal D}$, the posterior distribution of a probabilistic model with factors $\{t_i(\w)\}$ %_{i=1,\ldots,N}$
is
\begin{align} \label{eq:post}
p(\w|{\cal D}) = \frac{1}{Z}\prod_{0=1,\ldots,N} t_i(\w).
\end{align}
where $Z$ is the normalization constant. Note that the prior distribution over $\w$  is the factor $t_0$ in the above equation and  a factor $t_i(\w)$ may link to one, several, or all variables in $\w$.  In general, we do not have a closed-form solution for the posterior calculation. 
We could use random sampling methods---such as the Metropolis Hasting---to obtain the posterior distribution, but these methods can suffer on slow convergence, especially for high dimensional problems.

To reduce the computational cost, \citet{Minka01Thesis} proposed EP to approximate the posterior distribution $p(\w|D)$ by $q(\w)$ via factor approximation:
\begin{equation}
q(\w) = \prod_i \tt_i(\w)
\end{equation}
where $\tt_i(\w)$ approximates $t_i(\w)$ and has a simpler tractable form. EP requires both $q(\w)$ and the approximation factor $\tt_i(\w)$ have the form of the exponential family---such as Gaussian or factorized (or some structured) discrete distributions. The approximation factors are unnormalized, but given them, we can also easily obtain the natural parameters of the approximate posterior $q(\w)$ due to the log linear property of the exponential family.
For a graphical model representation, we can interpret the approximation factor $\tt_i(\w)$ as a message from the $i^{th}$ exact factor  $t_i(\w)$ to the variables linked to it.

To find the approximate posterior $q$, after initializing all the messages as one, EP iteratively refines the messages by repeating the following three steps:
 message deletion, belief projection, and message update, on each factor. In the message deletion step, we compute the partial posterior $\qnoti(\w)$ by removing a message $\tilde{t}_i$  from the approximate posterior 
$\qold(\w)$:
$\qnoti(\w)\propto \qold(\w)/\tilde{t}_i(\w)$. In the  projection step, we minimize the KL divergence between $\hatpi(\w) \propto t_i(\w) \qnoti(\w)$ and the new approximate posterior $q(\w)$,
such that the information from each factor is incorporated into $q(\w)$.
Finally, the message $\tilde{t}_i$ is updated via
$\tilde{t}_i(\w) \propto q(\w)/\qnoti(\w)$.

Since $q(\w)$ is in the exponential family, it has the following form
\[
q(\w) \propto \exp(\bnu\Tr \bphi(\w))
\]
where  $\bphi(\w)$ are the features of the exponential family. Given this representation, the KL minimization in the key projection step is achieved by moment matching:
\begin{align}
\int \bphi(\w) \hatpi(\w) \d\w = \int \bphi(\w)q(\w) \d\w
\end{align}

This KL minimization distributed on each factor works very well, when the data is relatively clean and the approximate posterior $q$ is not too far from $\hatpi$. However, in practice, the presence of outliers can ruin the distributed KL minimization and leads to divergence of the algorithm.

\section{Relaxed Expectation Propagation}
\label{REP}
In this section, we first present the new relaxed expectation propagation framework, discuss the choice of relaxation factors, and then describe its primal energy functions. 

\subsection{The REP Algorithm}
To reduce the impact of outlier factors, 
we can introduce relaxation factor $r_i(w) \propto \exp(\Beta_i\Tr \bphi(w))$ into the KL divergence. And to avoid too much relaxation we use a $l_1$ penality over it:
\begin{align} \label{eq:rKL}
 KL_r(\hatpi r_i|| q r_i ) + c |\Beta_i|_1  \raisetag{0.5in}
\end{align}
over $q$ and $r_i$,
where $|\Beta_i|_1$ is the $l_1$ norm of $\Beta_i$, the weight $c$ controls how much relaxation we have, and the $KL_r$ divergence is defined for unnormalized distributions.

This replacement allows us to adaptively handle factors---whether it is an outlier or not, and accurately approximate the posterior distribution $p(\w|{\cal D})$ \eqref{eq:post} by $q(\w) \propto \prod_i \tt_i(\w)$.

With this relaxed KL divergence, we obtain the following REP algorithm:
\begin{enumerate}
\item Initialize $q(\w)$ as the prior $t_0(\w)$ (assuming  the prior is in the exponential family) and all the messages $\tt_i(\w)=1$ for $i=1,\ldots,N$.
\item Loop until convergence or reaching the maximal number of iterations.
\begin{itemize}
\item Loop over factor $i=1,\ldots,N$:
\begin{enumerate}
\item {\bf Message deletion:} Based on the current factor $\tt_i$ and $\qold$, calculate the partial belief
\[
\qnoti \propto  \qold(\w)/\tilde{t}_i(\w).
\]
\item {\bf Belief projection:}
Incorporate information from the exact factor $t_i$ into the new belief $q$ by minimizing the penalized KL:
\begin{align} \label{eq:rKL2}
\min_{r_i,q}  KL_r(t_i r_i\qnoti|| q r_i ) + c |\Beta_i|_1  
\end{align}
where $\hatpi(\w) = t_i(\w)r_i(\w)\qnoti(\w)$.

\item {\bf Message update:} Update the message based on the new belief:
\[
\tilde{t}_i(\w) \propto q(\w)/\qnoti(\w).
\]
\end{enumerate}
\end{itemize}
\end{enumerate}

Unlike EP, REP does not require strict moment matching between $\hatpi(\w) \propto t_i(\w) \qnoti(\w)$ and the new approximate posterior $q(\w)$. How close these moments are depends on how big $\Beta_i$ is in the $l_1$ penalized relaxation factor $r_i$.

\subsection{Choice of relaxation factors}
For the relaxation factors $r_i(\w) = \exp(\Beta_i \bphi(\w))$, we should parameterize $\Beta_i$
in a form to make the minimization of \eqref{eq:rKL2} easy. Clearly, there are many choices available for us. 
A convenient one is to set (part of) $\Beta_i$ to be a scaled version of the parameters of an old message $\tilde t_i$, 
which can damp the influence of outliers via relaxed moment matching, but it will not cause double-counting of factors.  
The reason is that $r_i$ appears in both sides of \eqref{eq:rKL2} and the new posterior $q$ does not include $r_i$.
With this choice, we can use moment matching to easily obtain an analytical solution for the product of $q$ and $r_i$,
greatly simplifying the joint optimization over $q$ and $r_i$. This makes the computational overhead of REP over EP negligible in practice.

If we choose a form of $r_i$ that makes the joint minimization over $r_i$ and (i.e., belief) $q$ expensive, we can still use a sequential minimization procedure: first minimize the penalized KL to obtain $r_i$ based on the current $q$; and then, based on the estimated relaxation factor, minimize the relaxed KL to obtain the new $q$.

\subsection{Energy function}
Now we give the primal and dual energy functions for relaxed expectation propagation.
The primal energy function is
\begin{align}
 \min_{\Beta_i,{\hat p_i}}\max_{q} \sum_i \frac{1}{\hat{Z}_i}\int_\w \hat{p}_i(\w) r_i(\w) \log\frac{\hat{p}_i(\w)}{\hat{Z}_i t_i(\w)p(\w)} \nonumber \\
-(n-1)\frac{1}{Z_q} \int_{\w} q(\w) r_i(\w) \log \frac{q(\w)}{Z_q p(\w)} + c\sum_i|\Beta_i| \label{eq:PENG}
\end{align}
subject to
\begin{align}
\frac{1}{\hat{Z}_i}\int_{\w} \phi(\w)  \hat{p}_i(\w) r_i(\w) d \w &= \frac{1}{{Z}_q}\int_{\w} \phi(\w) q(\w) r_i(\w) d \w  \label{eq:mm}
\end{align}
where $\int_{\w} \hat{p}_i(\w) d \w      = 1$,
$\int_{\w} q(\w) d \w = 1$,
$\hat{Z}_i = \int_\w \hat{p}_i(\w) r_i(\w) d\w $,
and $Z_q = \int_\w q(\w) r_i(\w) d\w $.

Based on the KL duality bound, we obtain the dual form of the energy function (See the Appendix for details). Setting the gradient of the dual function to zero gives us the fixed-point updates described in the previous section. The fixed-point updates, however, do not guarantee convergence, just like the classical EP updates. However, REP is much more robust than EP;  in our experiments while EP diverges on difficult datasets, REP does {\em not} diverge in our experiments once. 

We believe the robustness of REP comes from the relaxation of moment matching in \eqref{eq:mm}: it does not demand the moments of $\hat{p}_i$ and $q$ to be exactly matched as in EP. Given an outlier factor, the exact moment matching requires the current $q$ moves dramatically to a new $q$, ignoring all the information from the previous factors, summarized in the current $q$. And this can cause oscillations, reducing the final approximation accuracy.

From an optimization perspective, the min-max cost function (\ref{eq:PENG}) includes the cost function of EP as a special case by setting $r_i(w)=1$. By tuning $r_i(w)$, it is possible to find a better solution to the min-max optimization. As shown by  \cite{Heskes:2005sf}, the cost function of EP corresponds to the Bethe energy, an entropy approximation, with exact moment matching constraints. 
With relaxed moment matching, we can potentially obtain better entropy approximation (We will further our research along this line in the future).

Finally we want to stress that by REP robustifies EP to obtain an more accurate posterior approximation, rather than ignoring information from outliers, as shown in figure~\ref{fig:toyexample}.

\section{REP training for Gaussian process classification}
In this section,  we present a REP-based training algorithm for Gaussian process classification.
First, let us denote $N$ independent and identically distributed samples as
$\mathcal{D}=\{(\x_i,y_i)\}_{i=1}^N$, where $\x_i$ is a  $d$ dimensional input and $y_i$ is a scalar output. We assume there is a latent function $f$ that we are modeling and the noisy realization of latent function $f$ at $\x_i$ is $y_i$.

We use a GP prior with zero mean over the latent function $f$. Its projection at the samples $\{\x_i\}$ defines a joint Gaussian distribution:
$p(\f) = \N(\f|\0, K)$
where 
$K_{ij}=k(\x_i,\x_j)$ is the covariance function, which encodes the prior notation of smoothness.
For classification, the data likelihood has the following form
\begin{align}\label{eq:clslik}
p(y_i|f)= (1-\epsilon) \Theta(f(\x_i) y_i) + \epsilon \Theta( - f(\x_i) y_i)
\end{align}
where $\epsilon$ models the labeling error, and $\Theta(a) = 1$ when $a\geq0$ ($\Theta(a) = 0$ otherwise).

Given the GP prior over $f$ and the data likelihood, the posterior process is
\begin{align}\label{eq:exactGP}
p(f|\mathcal{D}) \propto GP(f|0,K) \prod_{i=1}^{N} p(y_i|f(\x_i))
\end{align}
Due to the nonlinearity in $p(y_i|f)$, the posterior process does not have a closed-form solution.

Using REP, we approximate each non Gaussian factor $p(y_i|f(\x_i))$ by a Gaussian factor $\tt_i(f_i) = \N(f_i|m_i,v_i)$. Then we obtain a Gaussian process approximation to \eqref{eq:exactGP}:
\begin{align}\label{eq:approxgp}
p(f|\mathcal{D},\t) \propto GP(f|0,K) \prod_{i=1}^{N} \N(f_i|m_i,v_i)
\end{align}

We parameterize the relaxation factor $r_i$ as an Gaussian:
\begin{align}\label{eq:rgp}
r_i(f_i) \propto \N(f_i|m_i,b_i),
\end{align}
so that $r_i$ share the mean as $\tt_i$ and $b_i$ is the only free parameter in $r_i$.
For the convenience of the following presentation, we define $\tt_{i,b}(f_i) \equiv \N(f_i|\mbi, \vbi) \propto  r_i(f_i) \tt_i(f_i)$.
Now we give the relaxed EP algorithm for training a GP classifier.
\begin{enumerate}
	\item Initialize $m_i=0$, $v_i=\infty$, and $b_i=0$ for $\tilde{t}_i$. Also, initialize $r_i$, $h_i=0$,  $\A=\K$, and $\lambda_i=\K_{ii}$.
	\item Until all $(m_i, v_i, b_i)$ converge:
		  Loop $i=1,\ldots,N$:
	\begin{enumerate}
		\item Remove $\tilde{t}_i$ from the approximated posterior:
			\begin{align}
          \lambda_i^{\noti} & =(\frac{1}{\A_{ii}}-\frac{1}{v_i})^{-1}  & %\\			
          h_i^{\noti} &= h_i+\lambda_i^{\noti} v_i^{-1}(h_i-m_i)
			\end{align}
		\item Minimize the relaxed KL divergence over $b_i$ (i.e., $r_i$) by line search (See the Appendix).
		\item Multiple $\qnoti$ with $r_i$:
			\begin{align}
            \lambdaib &= 1/(1/\lambda_i^{\noti} + b_i)  &%\\
            \hib &= h_i^{\noti} - \lambdaib b_i (h_i^{\noti} - m_i)
            \end{align}
        \item Minimize the relaxed KL divergence to obtain $\tt_{i,b}$:
        	\begin{align}
            % z & = \frac{\hib}{\lambdaib}
            \alpha & = \frac{1}{\sqrt{\lambdaib}} \frac{(1-2\epsilon)\N(z|0,1)}{\epsilon+(1-2\epsilon)\psi(z)} \label{eq:alpha} & %\\
            \tilde{h}_i & = \hib +  \lambdaib \alpha \\
            \vbi & = \lambdaib (\frac{1}{\alpha_i \tilde{h}_i} -1) & %\\
            \mbi & =  \tilde{h}_i + \vbi \alpha
        	\end{align}
        where $z= \hib/\sqrt{\lambdaib}$ and $\psi(\cdot)$ is the standard normal cumulative density distribution.
                     
     \item Remove $r_i$ from $\tt_{i,b}$ to obtain $\tt_i$:
        \begin{align}
         % vi_temp = v_i;
         v_i & = 1/(1/\vbi + b_i) &%\quad %\\
         m_i & = v_i (\mbi/\vbi + m_i^{\textrm{old}} b_i)
        \end{align}
        \item Update $\A$ and $h_i$:
        %{\small
         \begin{align}
              \A & = \A - \frac{\a_i \a_i\Tr}{\delta + \A_{i,i}} &
             h_i  &=\sum_j \A_{ij} \frac{m_j}{v_j}
			\end{align}		
         %}
where   $\delta  = 1/(1/v_i - 1/v_i^{\textrm{old}})$ and $\a_i$ is the i-th column of $\A$.
		\end{enumerate}
\end{enumerate}
We will release our software implementation upon the publication.

\section{Related works}

\citet{minka-divergnece}  proposed Power EP (PEP) 
via the use of the $\alpha$-divergence \citep{Zhu:1995dk}. The framework includes  EP, fractional Belief propagation \citep{Wim:2002}, and variational Bayes as special cases, each of which is associated with a particular value $\alpha$ in the  $\alpha$ divergence.
In the presence of outliers, by using a power smaller than one for factors,  Power EP increases the algorithmic stability over EP. But it also changes the divergence used for minimization to an $\alpha$-divergence that is different from KL, the desired divergence for many problems (e.g,. classification). In contrast, REP adaptively relaxes the KL minimization for individual factors only when it becomes necessary.

We can damp the step size for message updates to help convergence, as suggested in \citep{minka:2004}. But for difficult cases, we need to use a very small step size, greatly reducing the convergence speed. Furthermore, damping does not guarantee convergence. As a result, without using any stepsize, our approach is a good alternative to fix EP for difficult cases.

\section{Experiments}\label{sec:exp}
In this section, we compare EP, PEP, and REP on approximation accuracy, convergence speed, and prediction accuracy for
on Gaussian process classification.  We chose GP classification as the test bed because EP has shown to be an excellent choice for approximation inference with GP classification models \citep{Kuss05}. 
For EP, we used the updates described in Chapter 5.4 of the Thesis of \citet{Minka01Thesis}. Since there is no previous work that uses PEP for training GP, we derived the updates and described them in the Appendix. The reason we compared REP with PEP is because PEP can also help stabilize EP, possibly improving the approximation quality.

\subsection{ Evaluation of posterior approximation accuracy}

\begin{figure*}[t!]
\vspace{.0in}
\begin{centering}
\hspace*{\fill}
\subfigure[{ Decision Boundaries}]{\includegraphics[width=0.35\linewidth]{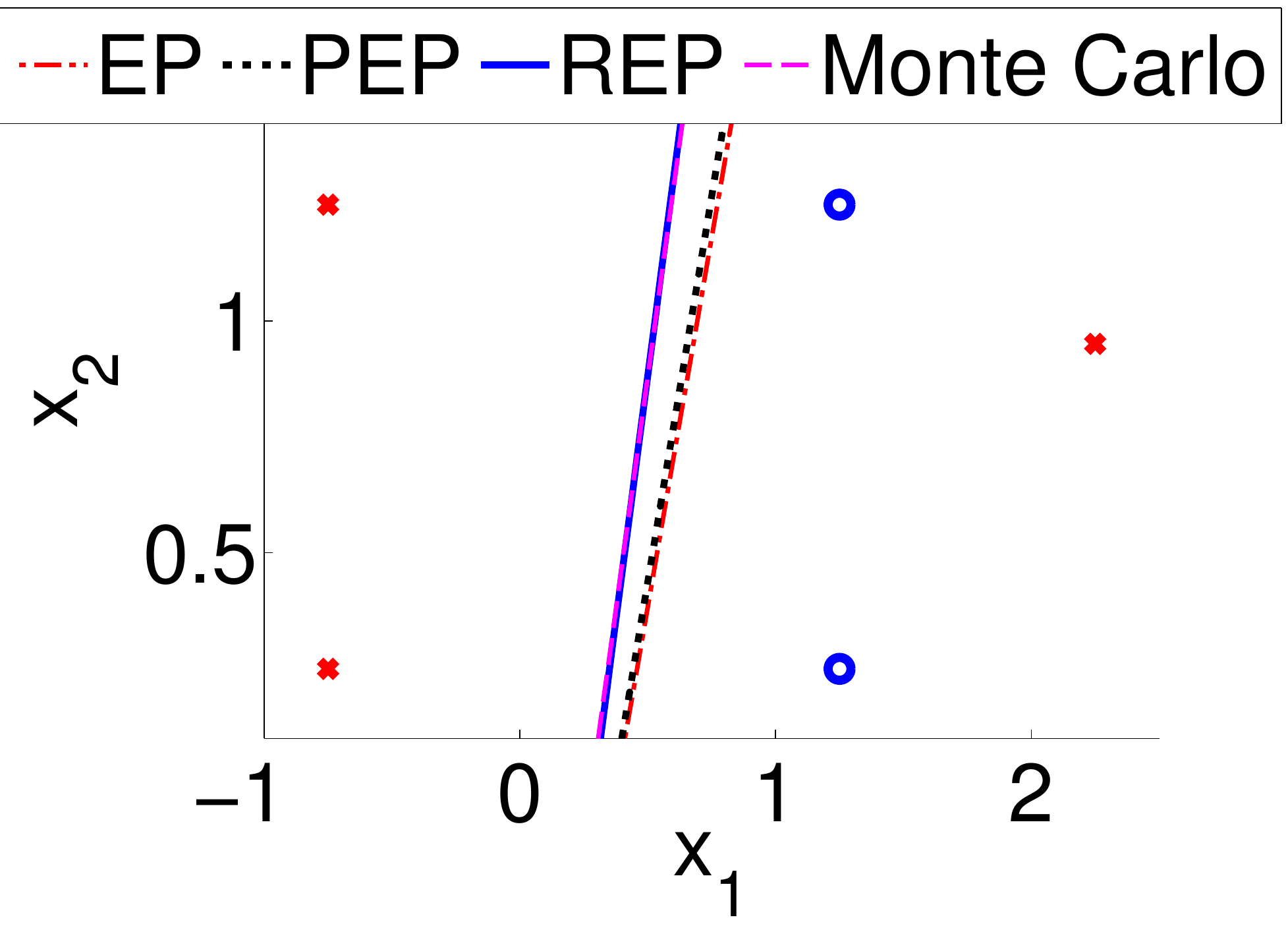}}
\hspace*{\fill}

\hspace*{\fill}
\subfigure[{ Error in mean est.}]{\includegraphics[width=0.35\linewidth]{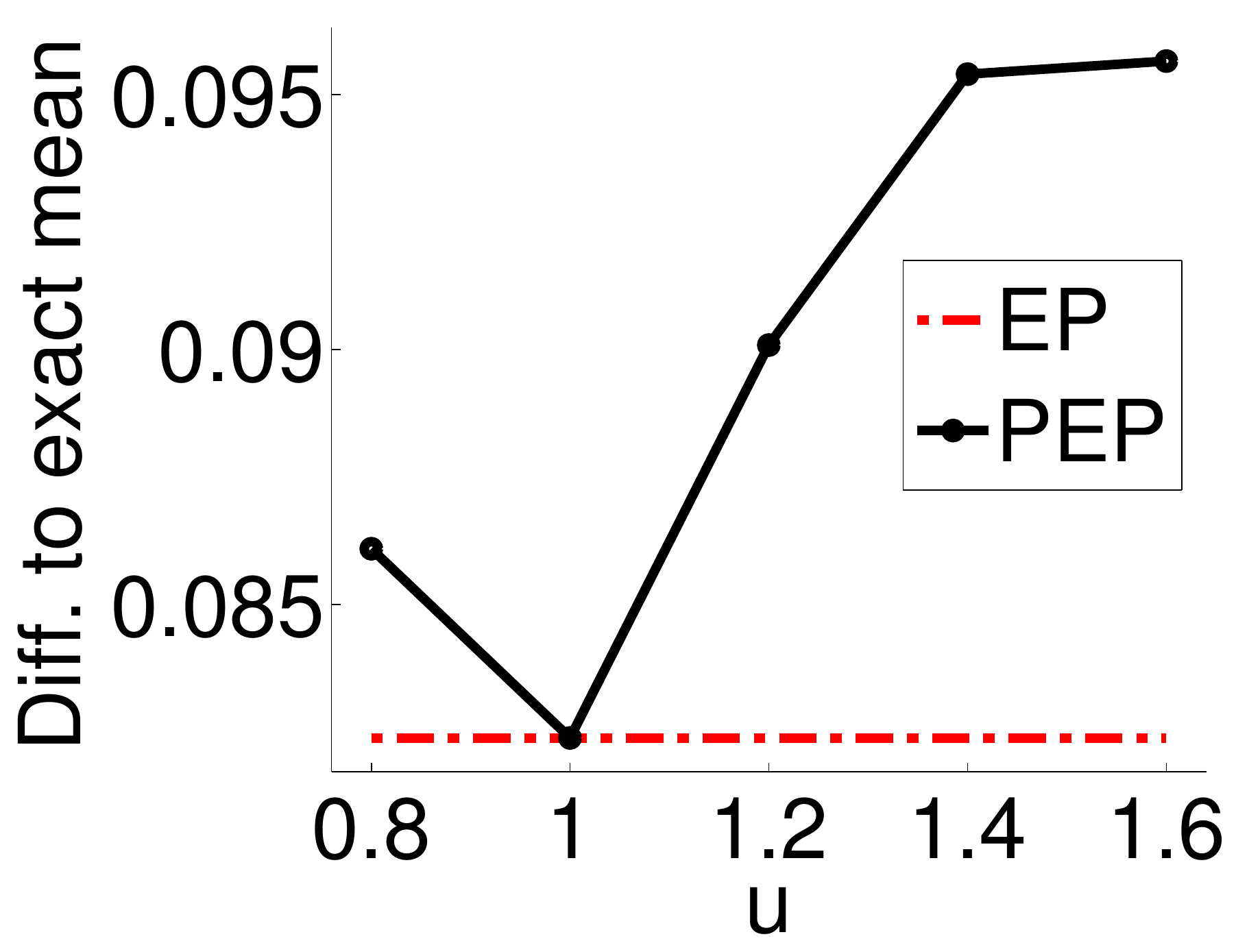}}
\hspace*{\fill}
\subfigure[{ Error in var. est.}]{\includegraphics[width=0.35\linewidth]{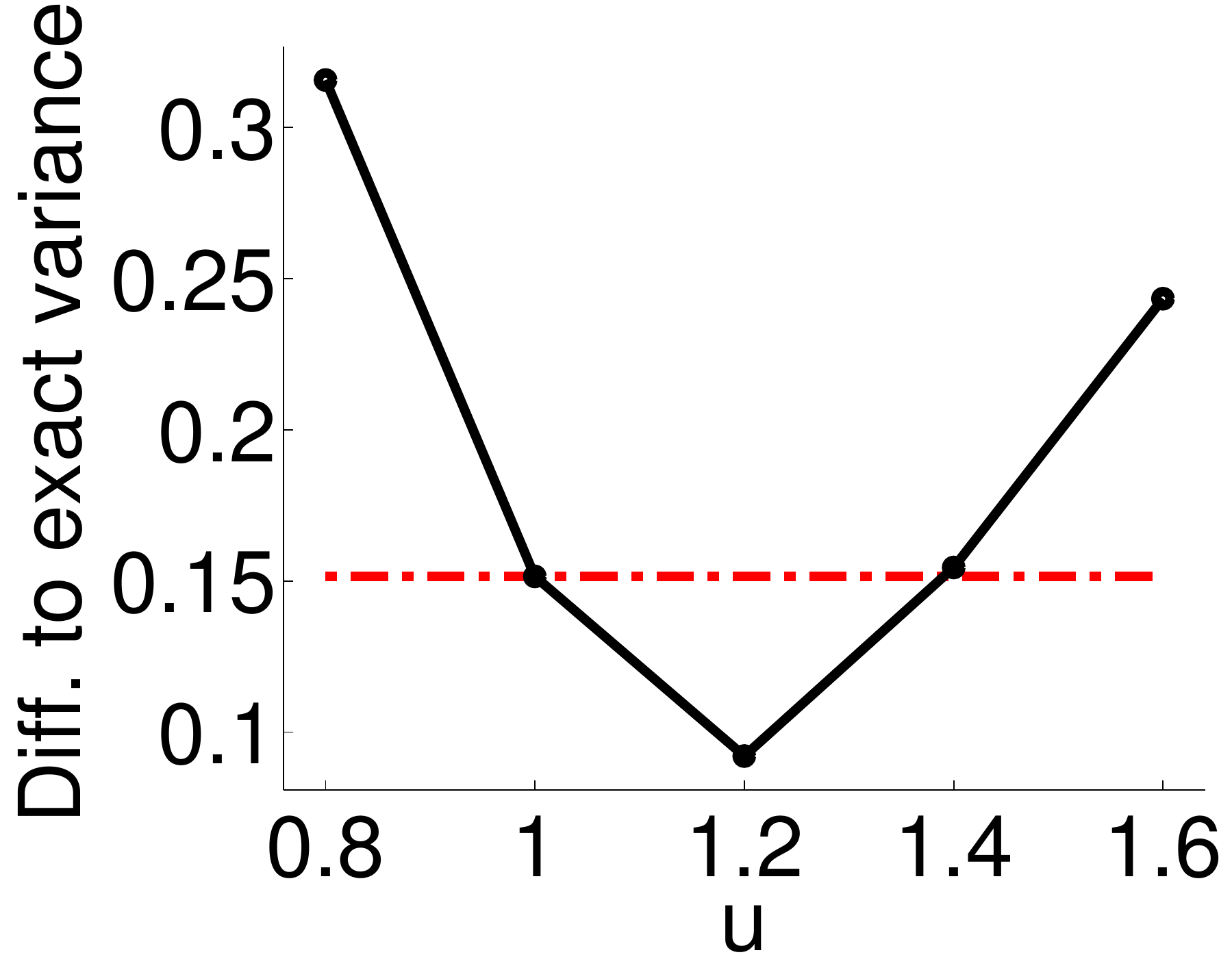}}
\hspace*{\fill}

\hspace*{\fill}
\subfigure[{ Error in mean est.}]{\includegraphics[width=0.35\linewidth]{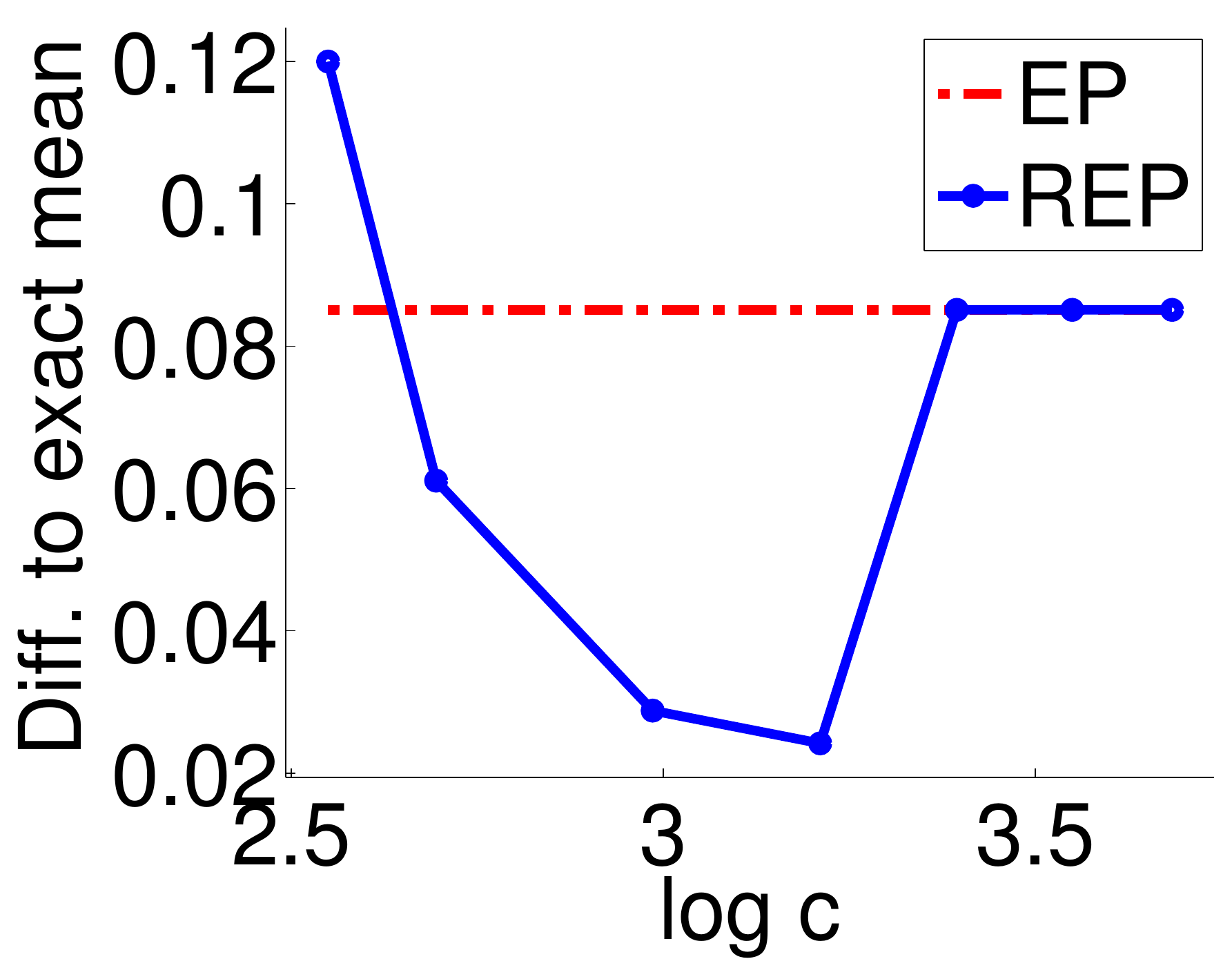}}
\hspace*{\fill}
\subfigure[{ Error in var. est.}]{\includegraphics[width=0.35\linewidth]{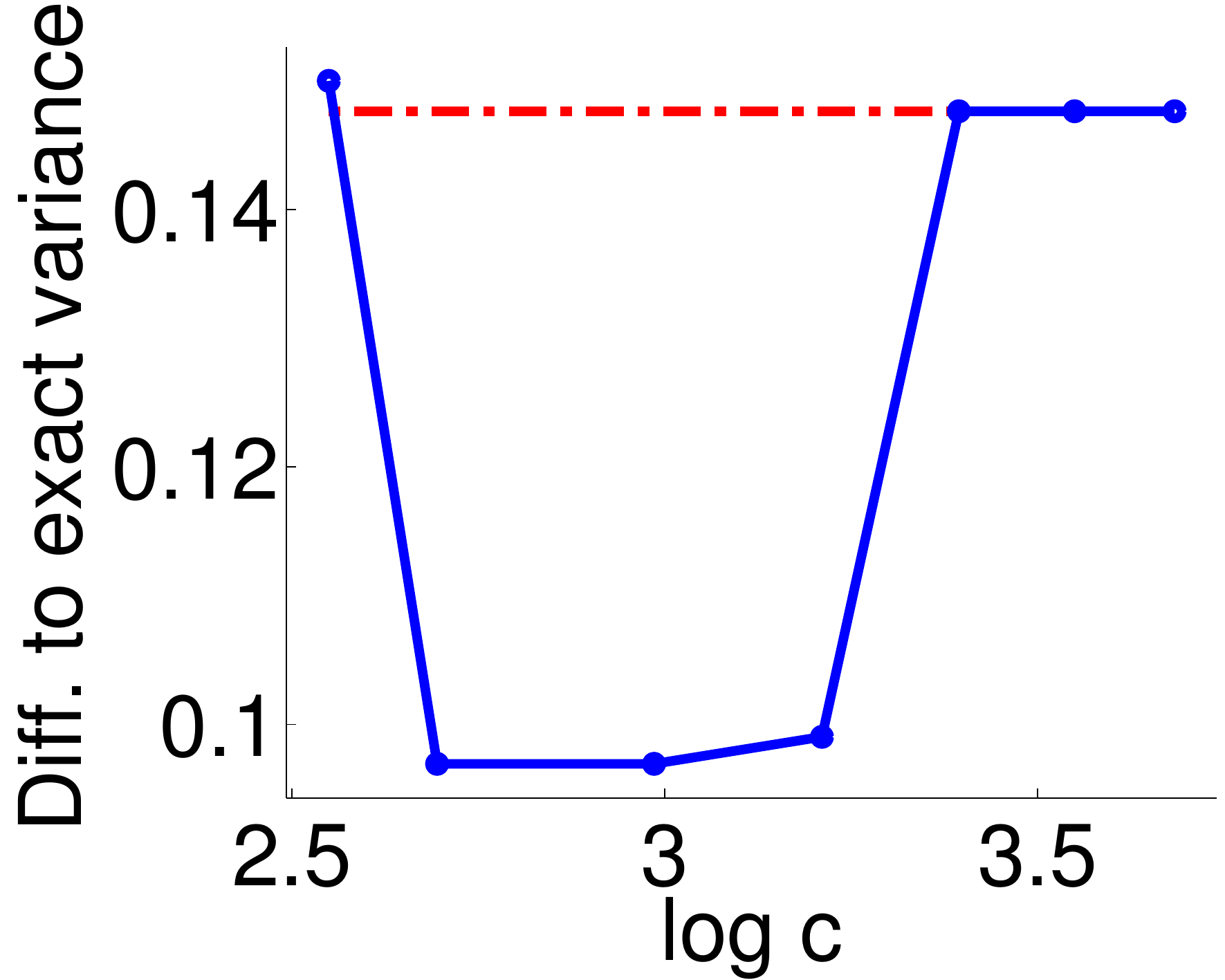}}
\hspace*{\fill}
\end{centering}
\vspace{.0in}
\caption{Classification of five data points, among which the red data point on the right is mislabeled.
(a): Decision boundaries of EP, Power EP, and Relaxed EP; (b) and (c): EP vs Power EP with different powers $u$;
(d) and (e): EP vs Relaxed EP with different penalty weights $c$.
REP reduces to EP when $c$ is big. For a wide range of $c$ values, the REP's approximation accuracy is significantly higher than those of EP and Power EP.
\alanc{ ($\alpha=0.5$ )} }
\label{fig:toyexample}
\end{figure*}

First, we considered linear classification of five data points shown in Figure \ref{fig:toyexample}.  The red `x' and blue `o' data points belong two classes. The red point on the right is mislabeled.
To reflect the true labeling error rate in the data, we set $\epsilon=0.2$ in \eqref{eq:clslik}. To obtain linear classifiers, we use the linear kernel---$k(\x_i,\x_j)=\x_i\Tr\x_j$ for the three algorithms.
After the algorithms converge, we can obtain the posterior mean and covariance of a linear classifier $\w$ in the 2-dimensional input space. 

To measure the approximation quality, we first used importance sampling with $10^8$ samples to obtain the exact posterior distribution of the classifier $\w$. We then applied these algorithms to obtain the approximate posteriors.
We treated the (approximate) posterior means as the estimated classifiers and used them to generate their decision boundaries. They are visualized  in Figure \ref{fig:toyexample}.a. For PEP, we set the power $u$ to 0.8; for REP, we set $c=20$.
Given the outlier on the right, the EP decision boundary significantly differs from the exact Bayesian decision boundary obtained from the importance sampling; the PEP decision boundary is closer to the exact one; and the REP decision boundary overlaps with the exact one perfectly.

We also varied the relaxation weight $c$ in \eqref{eq:rKL2} for REP and the power for PEP to examine their impact on approximate quality. We measured the mean square distances between the estimated and the exact mean vectors; we also computed the mean square distances between the estimated and the exact covariance matrices. The results are summarized in Figure \ref{fig:toyexample}.b to \ref{fig:toyexample}.e.
For PEP, as shown in  \ref{fig:toyexample}.b to \ref{fig:toyexample}.c, although the decision boundaries appear to be more aligned with the exact posterior distribution, their estimated mean and covariance are always worse than what EP achieve. This suggests that although PEP does reduce the influence of the outlier, it does not provide better approximation. By contrast, for REP,
when $c$ is big, the $l_1$ penalty forces the relaxation factor $b_i=0$ (i.e., $r_i =1$) and, accordingly, REP reduces to EP and gives the same results; And when  $c$ is relatively smaller (for a wide range of values), REP not only is immune to the presence of the outlier, but also improves the the approximation quality significantly.

Finally, we emphasize that REP aims to provide an accurate posterior approximation, regardless of likelihoods we used. For example, with various values of $\epsilon$ (e.g., 0.1 and 0.25) in \eqref{eq:clslik}, REP consistently provides more accurately results than EP and PEP.

\subsection{Results on synthetic data}

\begin{figure*}[th!]
\vspace{.0in}
\begin{centering}
\hspace*{\fill}
\subfigure[GP-EP: 80 iterations]{\includegraphics[width=0.45\linewidth]{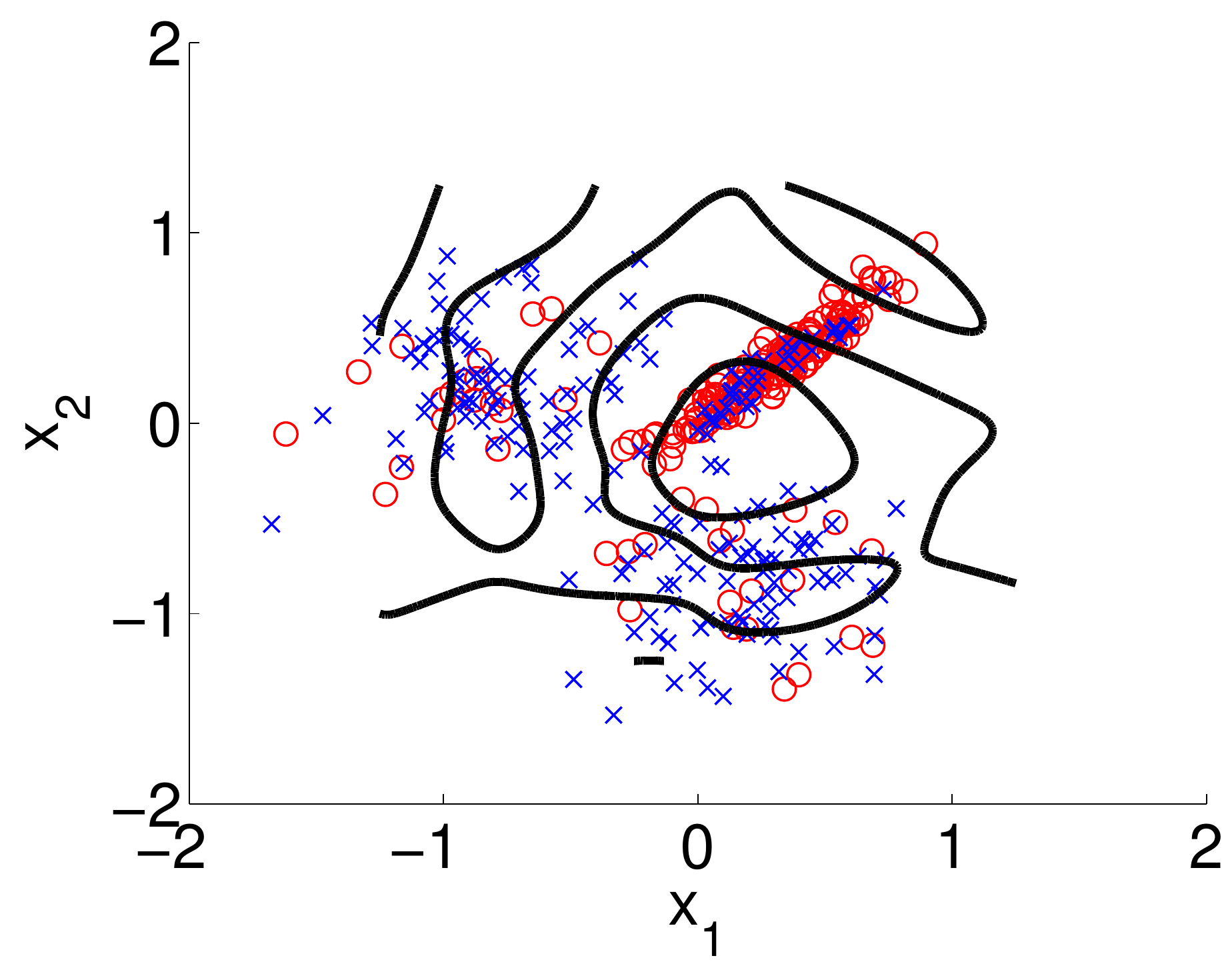}}
\hspace*{\fill}
\subfigure[GP-PEP: 20 iterations]{\includegraphics[width=0.45\linewidth]{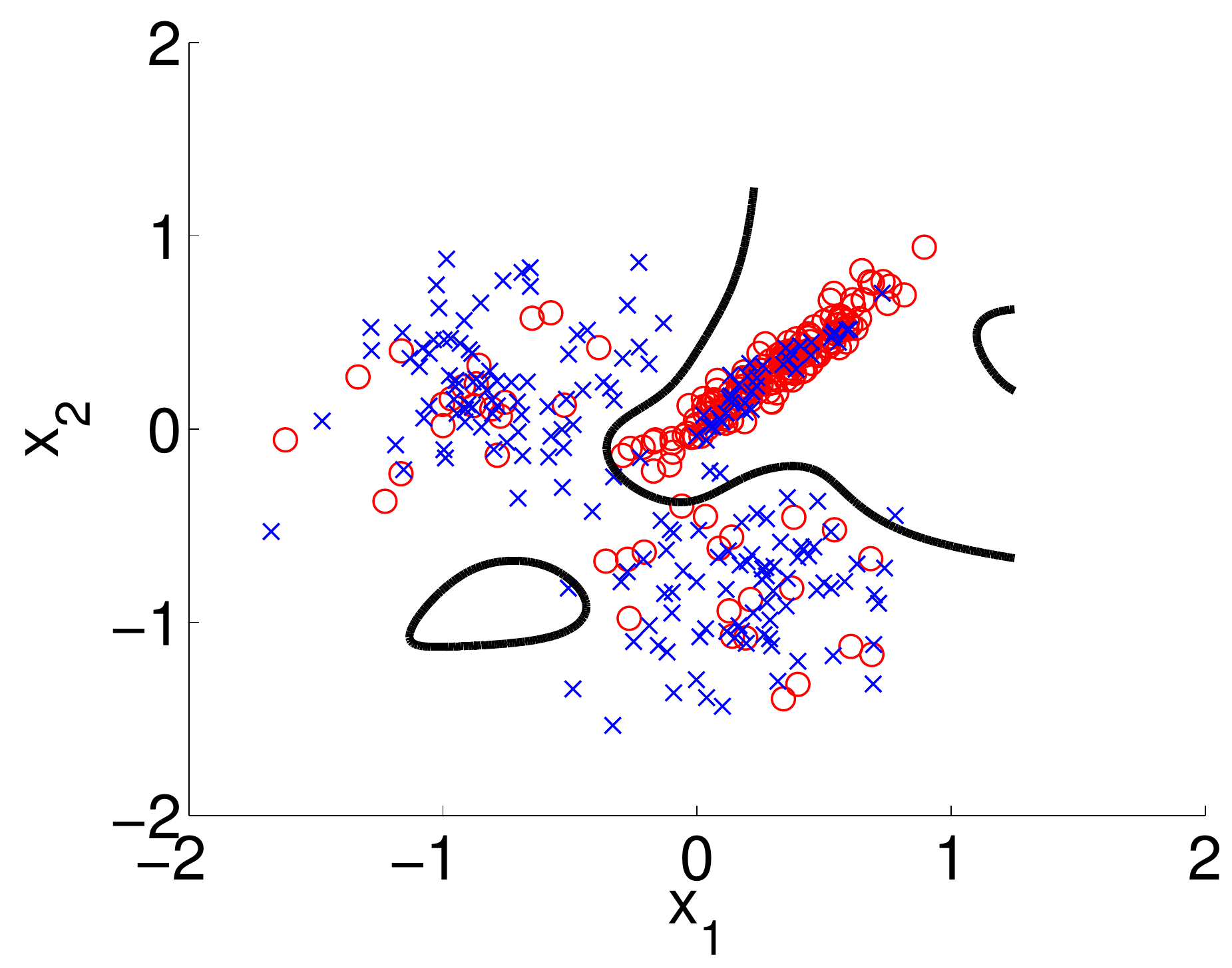}}
\hspace*{\fill}
\subfigure[GP-REP: 10 iterations]{\includegraphics[width=0.45\linewidth]{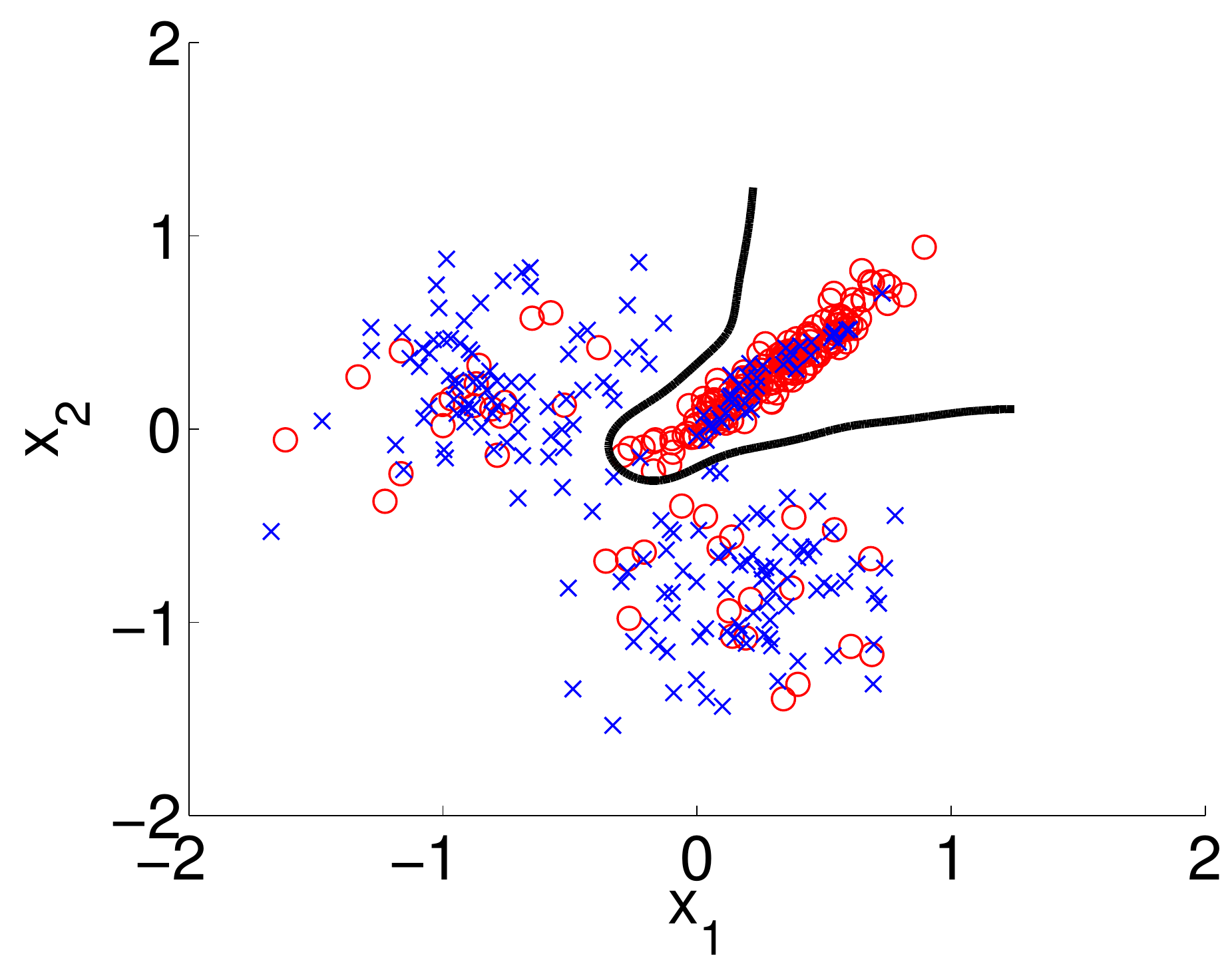}}
\hspace*{\fill}
\end{centering}
\vspace{.0in}
\caption{Decision boundaries of EP, Power EP, and REP. $20\%$ of the data points are mislabeled.}
\label{fig:Converge2}
\end{figure*}

 We then compared these algorithms on a nonlinear classification task. We sampled 200 data points for each class: for class 1 the points were sampled from a single Gaussian distribution and, for class 2, the points from a mixture of two Gaussian components. 
 The data points are represented by red crosses and blue circles for the two classes (See Figure \ref{fig:Converge2}).
We randomly flipped the labels of some data points to introduce labeling errors; we varied the error rates from $10\%$ to $20\%$. And for each case, we let $\epsilon$ match the error rate. We used a Gaussian kernel for all these training algorithms and applied cross-validation on the training data to tune the kernel width. We also tuned the relaxation weight $c$ for REP and the power for PEP.

\begin{figure}[ht]
\vspace{.0in}
\hspace*{\fill}
\subfigure[$10\%$ labeling error]{\includegraphics[width=0.45\linewidth]{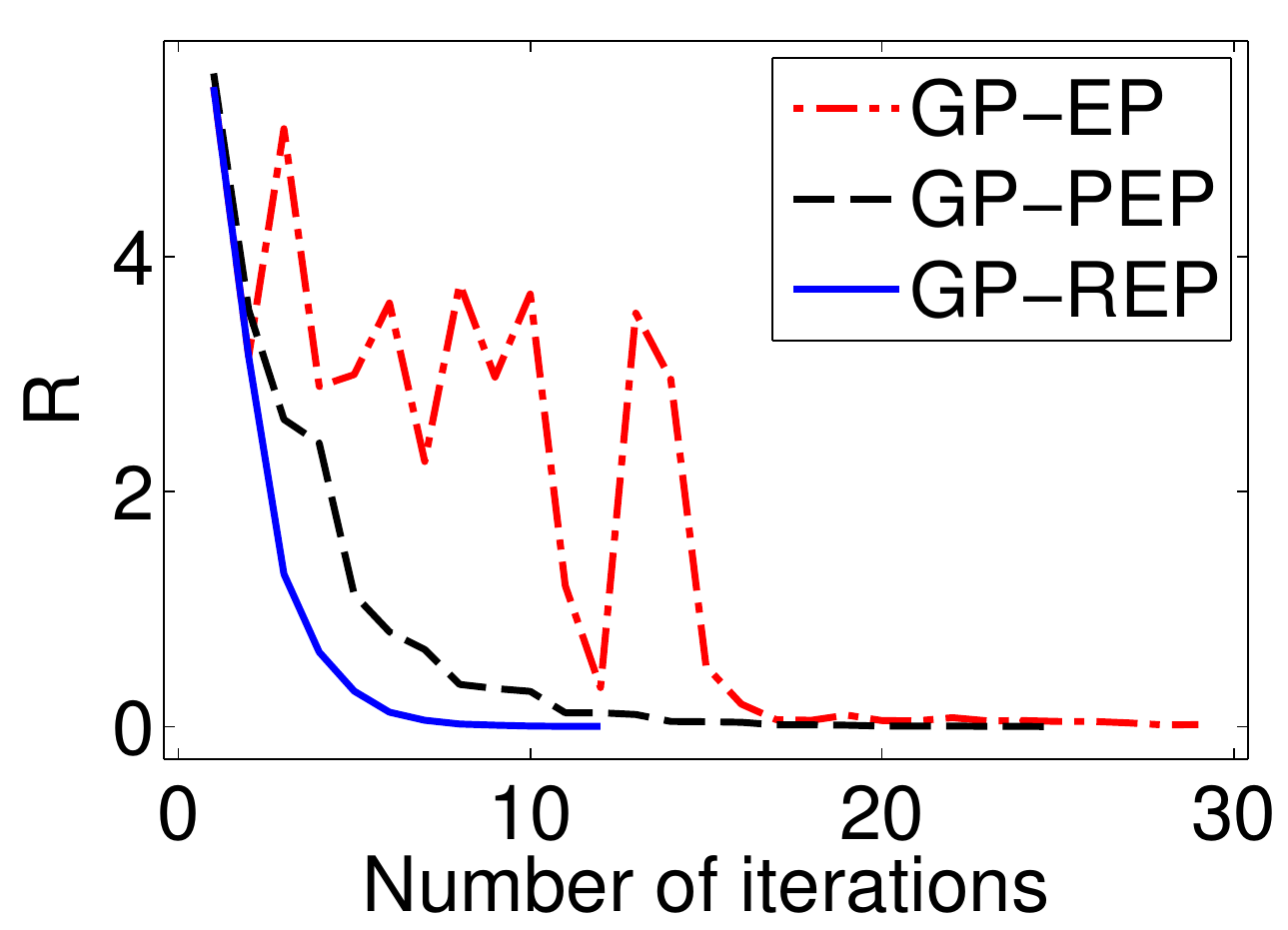}}
\hspace*{\fill}
\subfigure[$20\%$ labeling error]{\includegraphics[width=0.45\linewidth]{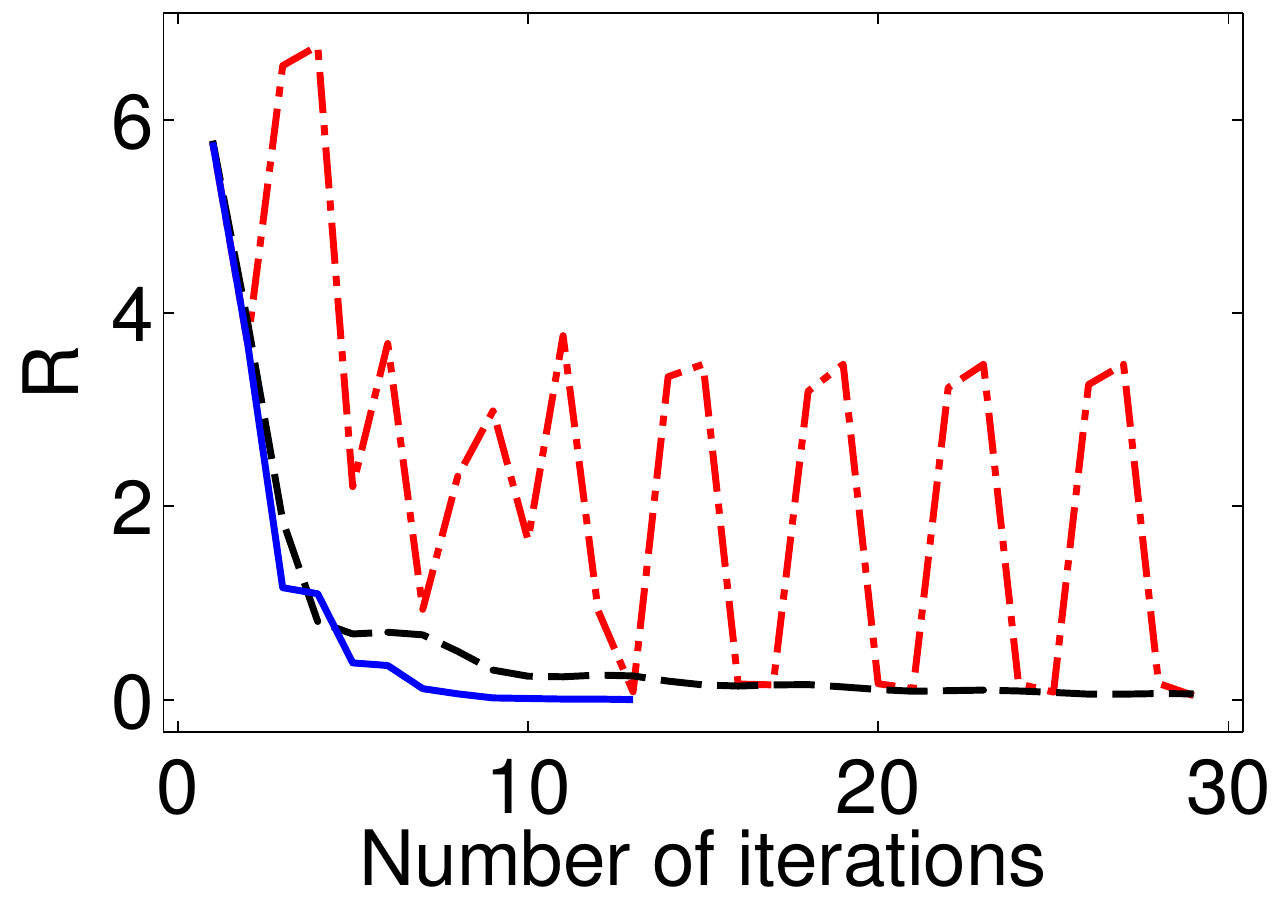}}
\hspace*{\fill}
\vspace{.0in}
\caption{Change in GP parameters along iterations.
}
\label{fig:R}
\end{figure}

\begin{figure*}
\begin{centering}
\vspace{.3in}
\hspace*{\fill}
\subfigure[Number of iterations]{\includegraphics[width=0.45\linewidth]{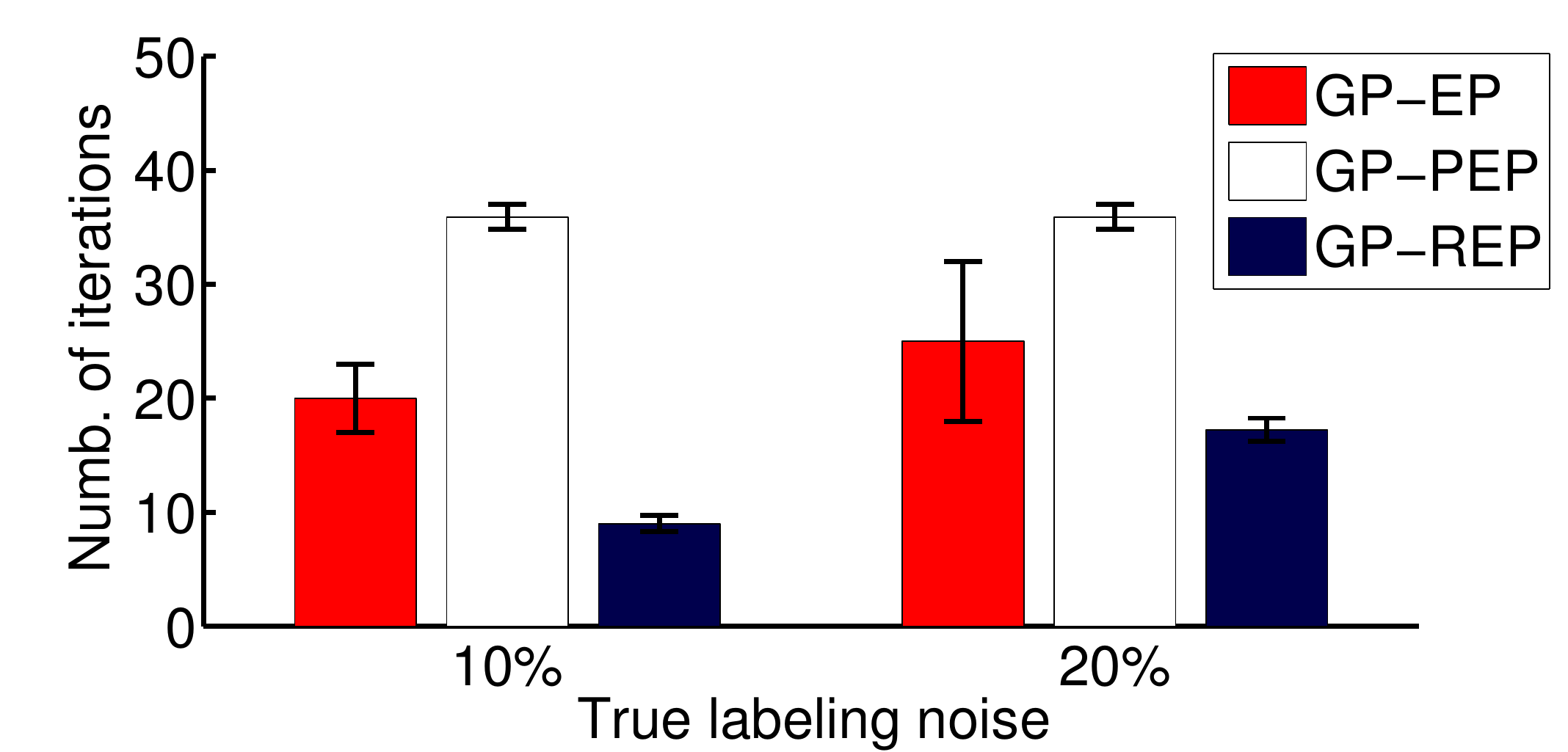}}
\hspace*{\fill}
\subfigure[Number of divergence]{\includegraphics[width=0.45\linewidth]{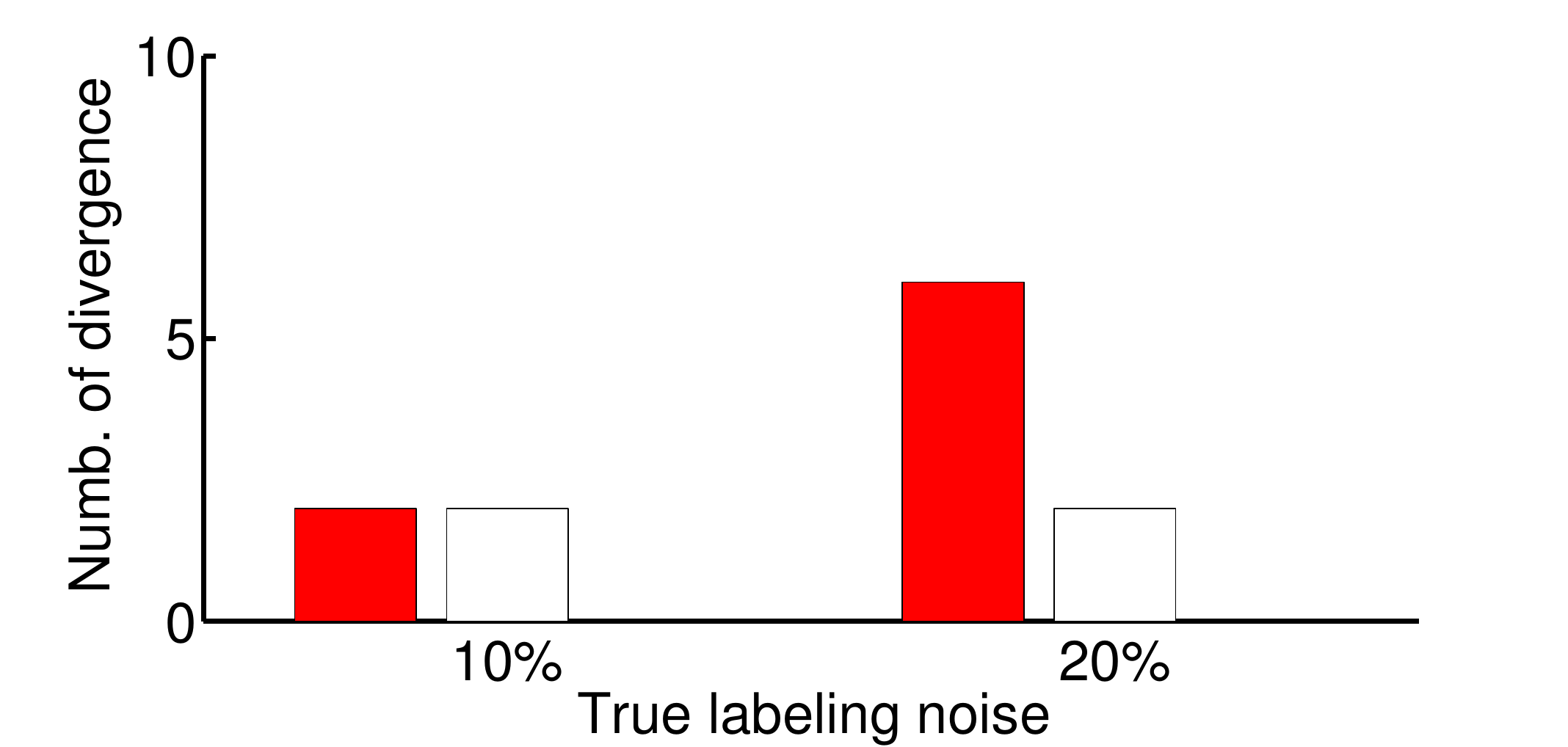}}
\hspace*{\fill}
\subfigure[Test error rate]{\includegraphics[width=0.45\linewidth]{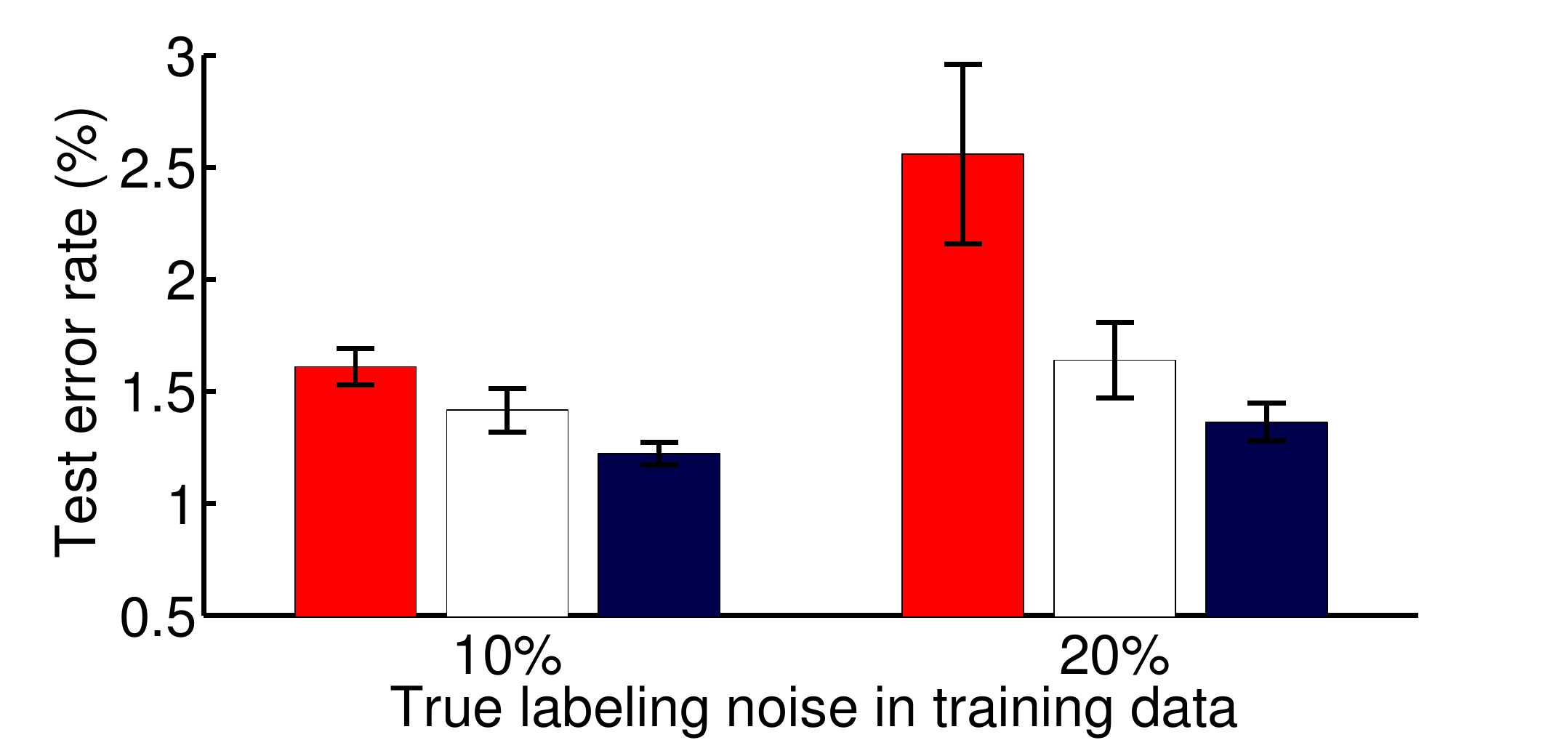}}
\hspace*{\fill}
\end{centering}
\vspace{-.1in}
\caption{Comparison of EP, Power EP, and Relaxed EP on two datasets with different labeling noise levels. Relaxed EP always converges. And with fewer iterations, Relaxed EP consistently achieves higher prediction accuracies than EP and Power EP. }
\label{fig:synres}
\end{figure*}

In Figure \ref{fig:Converge2}, we visualized the decision boundaries EP, PEP, and REP on one of the datasets with $20\%$ labeling errors.
To obtain these results, we set the power $u=0.8$ for Power EP and $c=10$ for Relaxed EP.
Clearly, EP diverges and leads to a chaotic decision boundary. PEP converges in 20 iterations and gives a decision boundary---better than that of EP but with strange shapes.  Finally, REP converges in only 10 iterations and provides a much more reasonable decision boundary than PEP.

To illustrate the convergence of  PEP and REP, we visualized in Figure \ref{fig:R} the change of the GP parameter $\balpha$ along iterations: $R(iter) \equiv \left\| \balpha_{iter}-\balpha_{iter-1}\right\|_2$. Clearly, PEP and REP are stabler than EP whose estimates oscillate---reflected by pikes in the $R$ curve.

We repeated the experiments 10 times; each time we sampled 400 training and 39,600 test points. Figure \ref{fig:synres} summarizes the results.
Figure \ref{fig:synres}.a shows that the number of iterations before convergence. The results are averaged over 10 runs.
To reach the convergence, we required $R<10^{-3}$. Clearly, REP converges faster than PEP and EP.

Figure \ref{fig:synres}.b shows that while EP and PEP can diverge (PEP diverges less frequently than EP), REP {\em always} converges.
Figure \ref{fig:synres}.c shows that REP gives significantly higher prediction accuracies than EP and PEP. Note that here we did not randomly flip the labels to introduce labeling errors in the test data and the prediction errors can be lower than the labeling errors in the training sets.

\subsection{Results on real data}

\begin{figure}
\vspace{.3in}
\hspace*{\fill}
\includegraphics[width=0.75\linewidth]{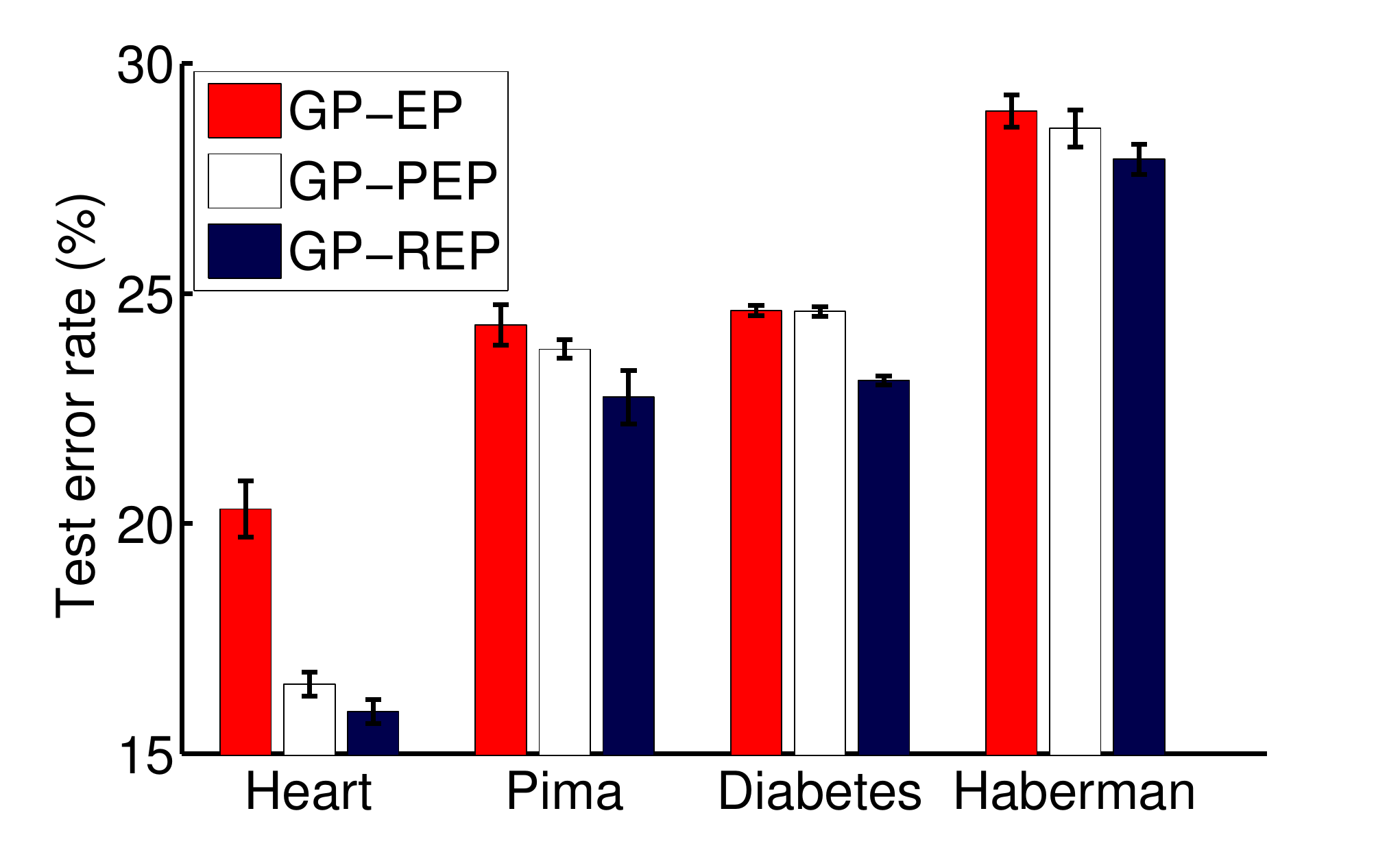}
\hspace*{\fill}
\vspace{.1in}
\caption{Test error rates of EP, PEP and REP on four UCI benchmark datasets without additional labeling noise.}
\label{fig:UCI}
\end{figure}

\begin{figure}
\vspace{.3in}
\hspace*{\fill}
\includegraphics[width=0.75\linewidth]{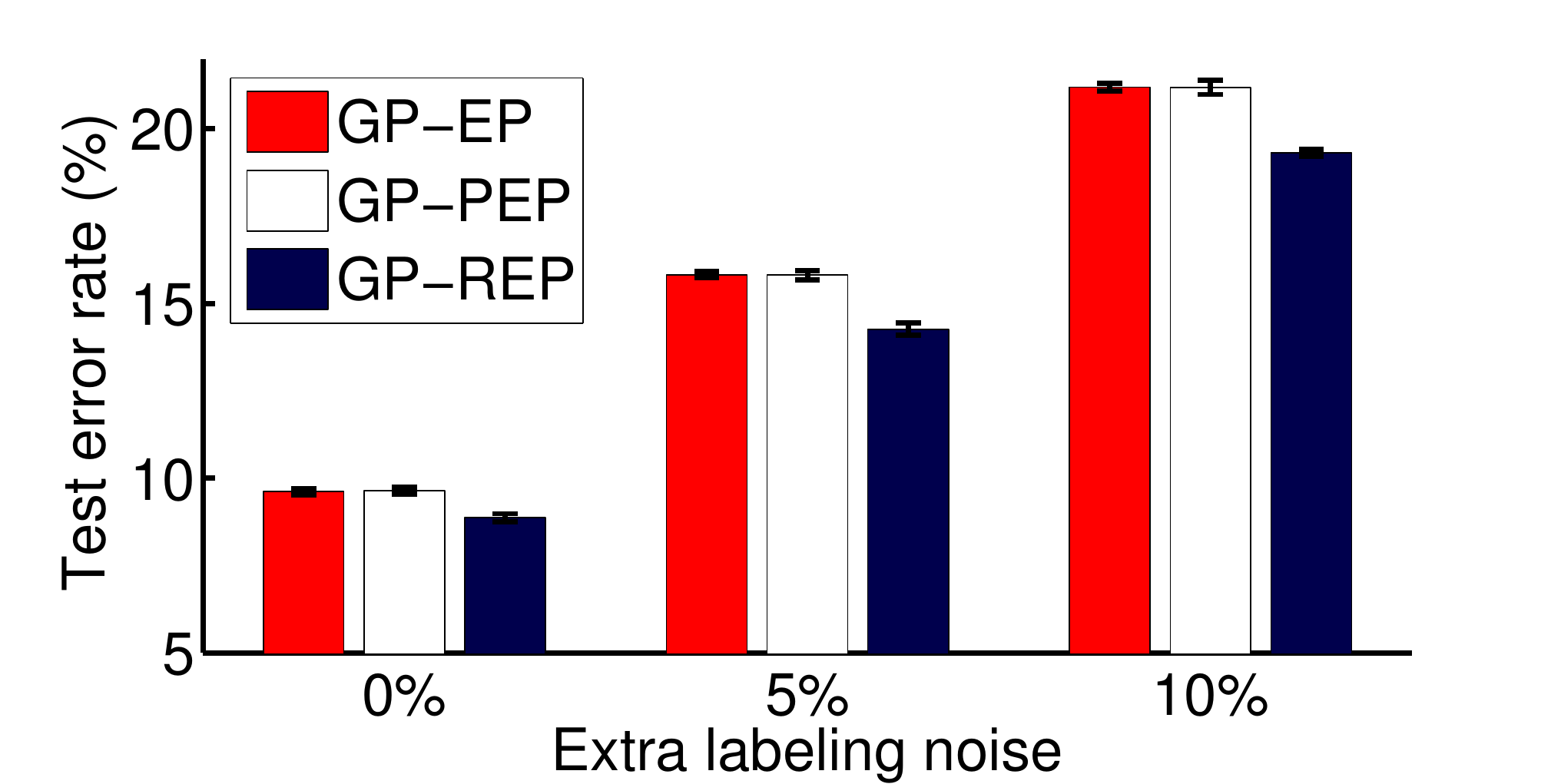}
\hspace*{\fill}
\vspace{.1in}
\caption{Test error rates of EP, PEP and REP on Spam dataset. We flipped the labels of some randomly selected data points to examine how these algorithms perform with outliers.}
\label{fig:RTR}
\end{figure}

Finally we tested these algorithms on five UCI benchmark datasets: Heart, Pima, Diabetes, Haberman, and Spam.

For the Heart dataset, the task is to detect heart diseases with $13$ features per sample. We randomly split the dataset into $81$ training and $189$ test samples 20 times. For the Pima dataset, we randomly split it into $319$ training and $213$ test samples, again 20 times. For the Diabetes dataset,
medical measurements and personal history are used to predict whether a patient is diabetic.
\citet{Ratsch:2001} split the UCI Diabetes dataset into two groups ($468$ training and $300$ test samples) for $100$ times. We used the same partitions in our experiments. For the Haberman's survival dataset, the task is to estimate whether the patient survive more than five years (including 5 years) after a surgery for breast cancer. The whole dataset contains information from $306$ patient samples and $3$ attributes per sample. We randomly split the dataset into $183$ training and $123$ test samples $100$ times.  Note that we did {\em not} add any labeling errors to these four datasets.
Figure \ref{fig:UCI} summarizes the results. The prediction accuracies of GP-EP and GP-REP are averaged over the splits of each dataset. REP outperforms the competing algorithms significantly.

For the Spam dataset, the task is to detect spam emails.  We partitioned the dataset to have $276$ training and $4325$ test samples, and flipped the labels of randomly selected data points from both the training and test samples. The experiment was repeated for $100$ times.
Figure \ref{fig:RTR} demonstrated that, with various additional labeling error rates, REP consistently achieves higher prediction accuracies than both EP and PEP.

\section{Conclusions}
In the paper we have introduced a method to increase the stability and approximation quality of EP.
We relax the moment matching requirement of EP with a $l_1$ penalty.  Experimental results on GP classification demonstrate that
 the new inference algorithm avoids divergence and gives higher prediction accuracy than EP and Power EP.

{
\bibliography{REP_2}
\bibliographystyle{unsrtnat}
}

% \small
%{
%\bibliography{REP_2}
%}
%\bibliographystyle{plainnat}
\begin{appendices}
%\appendix
\appendixpage
\section{Primal and dual energy functions for relaxed EP} 
The primary energy function of relaxed EP is
\begin{align}
\label{eq:APENG}
 \min_{\boldsymbol\eta_i}\min_{\hat p_i}\max_{q}  & \sum_i
\frac{1}{\hat{Z}_i}\int_\w \hat{p}_i(\w) r_i(\w) \log\frac{\hat{p}_i(\w)}{\hat{Z}_i t_i(\w)p(\w)}   \nonumber\\
&-(n-1)\frac{1}{Z_q} \int_{\w} q(\w) r_i(\w) \log \frac{q(\w)}{Z_q p(\w)}
+ c\sum_i|\boldsymbol\eta_i|_1
\end{align}
subject to
\begin{align}
\frac{1}{\hat{Z}_i}\int_{\w} \phi(\w)  \hat{p}_i(\w) r_i(\w) d \w &= \frac{1}{{Z}_q}\int_{\w} \phi(\w) q(\w) r_i(\w) d \w \\
\int_{\w} \hat{p}_i(\w) d \w        &= 1 \\
\int_{\w} q(\w) d \w                &= 1 \\
\hat{Z}_i                                 &= \int_\w \hat{p}_i(\w) r_i(\w) d\w \\
Z_q                                         &= \int_\w q(\w) r_i(\w) d\w \\
r_i(\w)                                    & \propto \exp(\boldsymbol\eta_i^T \phi(\w))
\end{align}
where $c$ is the constant and $r_i$ is the relaxation factor.

Based on the KL duality bound, we obtain the dual energy function. 
\begin{align}
&\min_{\boldsymbol\eta}\min_{\nu} \max_{\lambda}(n-1)\log \int_{\w}p(\w)\exp(\boldsymbol\nu^T \phi(\w)+\boldsymbol\eta_i^T \phi(\w)) d\w \nonumber\\
-&\sum_{i=1}^n\log \int_{\w} t_i(\w)p(\w)\exp(\boldsymbol\lambda_{i}^T \phi(\w)+\boldsymbol\eta_i^T \phi(\w)) d\w 
+ c \sum_i|\boldsymbol\eta_i|_1 \\
&(n-1)\boldsymbol\nu = \sum_i{\boldsymbol\lambda_{i}}
\label{eq:REngFun}
\end{align}
Setting the gradient of the above function to zero gives us the fixed-point updates described in the Section 3 of the main text. The fixed-point updates, however, do not guarantee convergence. But because of the relaxed KL minimization, REP always converges in our experiments (while EP can diverge when given many outliers).

Now we prove the duality of the relaxed EP energy function.
Applying the KL duality to the first term in (\ref{eq:PENG})produces
\begin{eqnarray}
\label{eq:RKLDUAL}
      &   &\frac{1}{\hat{Z}_i}\int_\w \hat{p}_i(\w) r_i(\w)  \log\frac{\hat{p}_i(\w)}{\hat{Z}_i t_i(\w)p(\w)} \\ \nonumber
 &=  &\frac{1}{\hat{Z}_i}\int_\w \hat{p}_i(\w) r_i(\w)  \log\frac{\hat{p}_i(\w)r_i(\w) }{\hat{Z}_i t_i(\w)p(\w)r_i(\w) } \\ \nonumber
 &=  &\max_{\lambda}\frac{1}{\hat{Z}_i} \int_{\w} \hat{p}_i(\w) r_i(\w) \boldsymbol\lambda_{i}(\w) d\w
 -\log \int_{\w} t_i(\w)p(\w)r_i(\w) \exp(\boldsymbol\lambda_{i}(\w))  d\w
\end{eqnarray}

This is because the maximum of the right side of (\ref{eq:RKLDUAL}) is achieved when
(taking derivative to $\boldsymbol\lambda_{i}(\w)$)
\begin{equation}
\frac{1}{\hat{Z}_i}\hat{p}_i(\w) r_i(\w)
- \frac{t_i(\w)p(\w) r_i(\w) \exp(\boldsymbol\lambda_{i}(\w))}{\int_{\w} t_i(\w)p(\w)r_i(\w)  \exp(\boldsymbol\lambda_{i}(\w)) d\w}
= 0
\end{equation}
which means
\begin{equation}
\exp(\boldsymbol\lambda_{i}(\w)) = \frac{\hat{p}_i(\w) r_i(\w) \int_{\w} t_i(\w) p(\w) r_i(\w) \exp(\boldsymbol\lambda_{i}(\w)) d\w }
{ \hat{Z}_i t_i(\w)p(\w)r_i(\w) }
\label{eq:RKLDUALD}
\end{equation}
Inserting $\exp(\boldsymbol\lambda_{i}(\w))$ in (\ref{eq:RKLDUAL})
proves the KL duality for (\ref{eq:RKLDUAL}). \\
And from the stationary condition, we can assume w.l.o.g. that
\begin{equation}
\label{eq:DecLam}
\boldsymbol\lambda_i(\w) = \boldsymbol\lambda_i^T\phi(\w)
\end{equation}
\begin{align}
       &\frac{1}{\hat{Z}_i}\int_\w \hat{p}_i(\w) r_i(\w)  \log\frac{\hat{p}_i(\w)}{t_i(\w)p(\w)}  \\\nonumber
  =  &\max_{\lambda}  \frac{1}{\hat{Z}_i}  \int_{\w} \hat{p}_i(\w) r_i(\w) \boldsymbol\lambda_{i}^T\phi(\w) d\w  -\log \int_{\w} t_i(\w)p(\w)r_i(\w) \exp(\boldsymbol\lambda_{i}^T\phi(\w))  d\w
\end{align}

Similarly, we have
\begin{align}
	&-\frac{1}{Z_q}\int_{\w} q(\w) r_i(\w) \log \frac{q(\w)}{Z_qp(\w)}  \\\nonumber
=       &-\frac{1}{Z_q}\int_{\w} q(\w) r_i(\w) \log \frac{q(\w)r_i(\w)}{Z_qp(\w)r_i(\w)}  \\\nonumber
=       &\min_{\boldsymbol\nu} - \frac{1}{Z_q}\int_{\w} \boldsymbol\nu(\w)q(\w)r_i(\w)d\w
+ \log \int_{\w} p(\w) r_i(\w) \exp( \boldsymbol\nu(\w)) d\w    \\\nonumber
=      &\min_{\boldsymbol\nu} - \frac{1}{Z_q}\int_{\w} \boldsymbol\nu^T\phi(\w)q(\w)r_i(\w)d\w
+ \log \int_{\w} p(\w) r_i(\w) \exp( \boldsymbol\nu^T\phi(\w)) d\w
\end{align}

With the constraint ($(n-1)\boldsymbol \nu = \sum_i \boldsymbol \lambda_i$) and (2),
we obtain the dual energy function:
\begin{align}
&\min_{\boldsymbol\eta}\min_{\boldsymbol\nu} \max_{\boldsymbol\lambda}(n-1)\log \int_{\w}p(\w) r_i(\w) \exp(\boldsymbol\nu^T \phi(\w)) d\w \nonumber\\
-&\sum_{i=1}^n\log \int_{\w} t_i(\w)p(\w) r_i(\w) \exp(\boldsymbol\lambda_{i}^T \phi(\w)) d\w  + c \sum_i|\boldsymbol\eta_i|_1  \\
&(n-1)\boldsymbol\nu = \sum_i{\boldsymbol\lambda_{i}}
\label{eq:REngCon}
\end{align}

\section{Relaxed KL for GP classification}

For GP classification, we minimize the relaxed KL divergence with $l_1$ penalty over $b_i$ by line search. Here we present how to compute the value
of this cost function:
\begin{equation}
\label{eq:GKLr}
Q(b_i) = KL_r(t_i r_i q^{\noti}  || r_i q)+ c |b_i|
\end{equation}

Following the notations in the main text (from equations (16) to (23)), we have $Q(b_i)$ as 
\begin{align}
& \frac{1}{ \hat{Z}_i} \left\{   \left[ (1-\epsilon) \log(1-\epsilon) - \epsilon \log\epsilon \right] \psi(z) + \epsilon\log\epsilon \right\}
+ \frac{1}{2 v_{i,b}} (F_{i,b} - \tilde{h_i} m_{i,b})  \nonumber \\
- &\frac{1}{2} \log \left(1 + (b_i + \frac{1}{v_{i,b}})\lambda^{\noti}_i \right) 
+\frac{1}{2}\log (b_i \lambda^{\noti}_i +1) - \frac{1}{2} b_i (m_i^2 -2 m_i \tilde{h_i} + F_{i,b}) \nonumber \\
+&\frac{1}{2} \frac{(m_i - h_i^{\noti})^2}{\lambda^{\noti}_i + b_i^{-1}} -\log \hat Z_i + c |b_i|
\end{align}
where $\hat{Z}_i  =  \epsilon + (1-2\epsilon) \psi(z)$,
and the term $F_{i,b}$ can be computed as follows:
\begin{align}
\delta_{i,b}  &  = (\frac{1}{v_{i,b}} - \frac{1}{v_i})^{-1} \\
a_{ii}^{new} &= (\frac{1}{a_{ii}} + \frac{1}{\delta})^{-1} \\
\tilde a_{ii}^{new} &= a_{ii}^{new} (1 - \frac{a_{ii}^{new}}{a_{ii}^{new} + b_i^{-1}})\\
F_{i,b}             &= \tilde a_{ii}^{new} + \tilde{h_i} ^2
 \end{align}

Using the above equations, we can efficiently optimize $Q(b_i)$ over $b_i$ via line search.

\section{Power EP for GP classification}

In this section, we describe how to train GP classifiers by Power EP.
The updates of Power EP are the same as equations  (5.64) to (5.74) in \citep{Minka01Thesis}, except two critical modifications:
\begin{itemize}
\item Replace equation (5.67) in \citep{Minka01Thesis} by
\begin{equation}
\alpha_i = \frac{1}{\sqrt{\lambda_i}}
\frac{ \left[ (1-\epsilon)^u - \epsilon^u \right] \N(z|0,1)}{ \epsilon^u + \left[(1-\epsilon)^u - \epsilon^u \right] \psi (z) }
\end{equation}
where $\psi(\cdot)$ is the standard normal cumulative density function and $u$ is the power used by Power EP.

\item
Moreover, after (5.70), scale $v_i$ by $u$:
\begin{equation}
v_i \leftarrow  u v_i
\end{equation}
\end{itemize}

%\section{References}
%[1] T. P. Minka. \textit{A family of algorithms for approximate Bayesian inference.}
%PhD thesis, Massachusetts Institute of Technology, 2011.

\end{appendices}
\end{document}